\documentclass[letterpaper, 10 pt, conference]{ieeeconf}  

\IEEEoverridecommandlockouts                              

\usepackage{amsmath} 
\usepackage{amssymb}  
\usepackage{todonotes}
\let\labelindent\relax 
\usepackage{enumitem}
\usepackage{mathtools}
\usepackage{rotating}
\usepackage{multirow}
\usepackage{siunitx}
\usepackage{caption}
\usepackage{subcaption}
\usepackage{float}
\usepackage{hyperref}
\usepackage{pifont}
\usepackage{cite}
\usepackage{microtype}

\usepackage{booktabs} 

\usepackage{algorithmic} 
\usepackage{algorithm2e}
\graphicspath{{graphics/}}

\usepackage{tabto} 

\hypersetup{
    colorlinks=true,
    linkcolor=black,
    filecolor=magenta,      
    urlcolor=cyan,
    pdftitle={Criticality Metrics for Object Detection},
    pdfpagemode=FullScreen,
    }

\title{\LARGE \bf
Criticality Metrics for Relevance Classification in Safety Evaluation of Object Detection in Automated Driving
}

\author{Jörg Gamerdinger, Sven Teufel, Stephan Amann, and Oliver Bringmann
\thanks{University of T\"ubingen, Faculty of Science, Department of Computer Science, Embedded Systems Group 
\tt\small {\{joerg.gamerdinger, sven.teufel, stephan.amann, oliver.bringmann\} @uni-tuebingen.de}}
}%

\begin{document}
\maketitle
\thispagestyle{empty}
\pagestyle{empty}

\begin{abstract}

Ensuring safety is the primary objective of automated driving, which necessitates a comprehensive and accurate perception of the environment. While numerous performance evaluation metrics exist for assessing perception capabilities, incorporating safety-specific metrics is essential to reliably evaluate object detection systems. A key component for safety evaluation is the ability to distinguish between relevant and non-relevant objects — a challenge addressed by criticality or relevance metrics. This paper presents the first in-depth analysis of criticality metrics for safety evaluation of object detection systems. Through a comprehensive review of existing literature, we identify and assess a range of applicable metrics. Their effectiveness is empirically validated using the DeepAccident dataset, which features a variety of safety-critical scenarios. To enhance evaluation accuracy, we propose two novel application strategies: bidirectional criticality rating and multi-metric aggregation. Our approach demonstrates up to a 100\,\% improvement in terms of criticality classification accuracy, highlighting its potential to significantly advance the safety evaluation of object detection systems in automated vehicles.

\end{abstract}

\section{INTRODUCTION}
\label{sec:intro}
\begin{figure}[t]
    \centering
    \includegraphics[width=\linewidth]{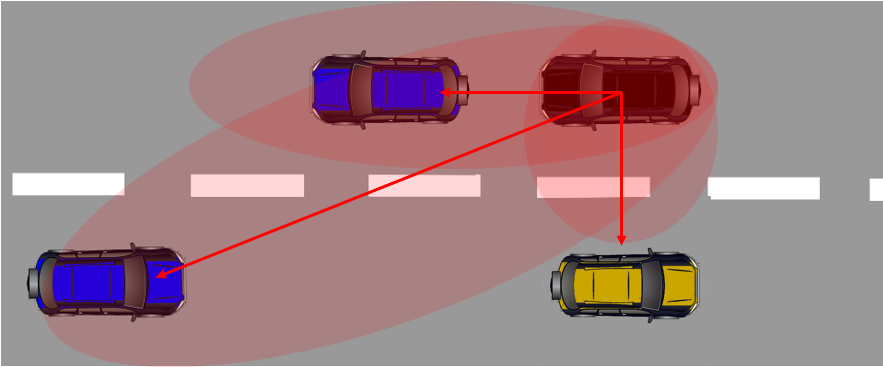}
    \caption{Exemplary Scene to represent criticality. The blue vehicles can be considered as critical or relevant as they could potentially collide with the ego (black). For the yellow vehicle a collision is unlikely; hence, the vehicle can be considered as not critical. Figure adapted from~\cite{volk2020safety}}
    \label{fig:criticality_example}
    \vspace*{-3mm}
\end{figure}

Automated driving is a promising approach to reducing road accidents and improving road safety. To achieve safe automated driving, automated vehicles (AVs) require a complete and correct perception of the environment. 
However, as shown by Volk et al.~\cite{volk2020safety}, performance metrics such as precision, recall or average precision (AP) are not suitable to determine whether an object detection system is sufficient to achieve safety. Therefore, appropriate metrics are required for the safety evaluation of object detection systems. A key component of safety evaluation is the relevance of an object in the environment. Relevance or criticality describes the need to perceive the object in order to avoid a safety-critical situation. 
For example, a vehicle in front of the ego vehicle needs to be perceived because it could lead to a collision if it isn't detected; therefore the vehicle is considered relevant or safety-critical. Considering a motorway scenario with infrastructure elements between driving directions, it is not necessary to perceive vehicles on the other side of the guardrail from a safety perspective.  An example scene with two critical objects (blue) for one ego (black) is shown in Fig.~\ref{fig:criticality_example}.

Current reviews on criticality metrics~\cite{mahmud2017application, westhofen2023criticality} focus only on providing an overview of metrics, but have not performed an evaluation using safety-critical scenarios. They also provide a more general overview. In this paper, we present the first technical review that identifies metrics that are applicable to the safety evaluation of object detection systems and evaluate them on a dataset containing safety-critical scenarios.

The main contributions of this work are:
\begin{itemize}
    \item We provide the first comprehensive review on criticality metrics which are suitable for offline safety evaluation of object detection systems
    \item We conduct an extensive evaluation using an accident dataset showing the limitations of state-of-the-art metrics
    \item We propose two novel application strategies: Multi-metric aggregation and bidirectional criticality rating to improve the state-of-the-art in terms of frame ratio by up to \SI{100}{\percent}
\end{itemize}

In Sec.~\ref{sec:metrics_review} we give an overview on current criticality metrics applicable to the safety assessment for object detection systems in automated driving. Section~\ref{sec:strategies} presents our proposed application strategies. A comprehensive evaluation and a discussion of the findings is performed in Sec.~\ref{sec:eval}. Finally, we conclude our work and give an outlook to further research.

\section{CRITICALITY METRICS}
\label{sec:metrics_review}
Criticality metrics can be applied to different tasks such as perception evaluation, scenario evaluation or motion planning. In this work, the focus is on metrics that can be applied to the safety evaluation of perception tasks.
Most of these metrics are based on vehicle dynamics information such as speed, safety margins and appropriate acceleration values. In the following, suitable metrics are presented in detail. 

For all metrics $v_i$, $a_i$, and $s_i$ describe the velocity, acceleration, and size of a road user $V_i$, where $V_1$ is the ego vehicle. $d_{1,2}$ represents the distance between $V_1$ and $V_2$. Size and distance are in \si{m}, velocity in \si{\metre\per\second} and acceleration in \si{\metre\per\second\squared}.

\subsection{Time-to-Collision}
\label{subsubsec:ttc}
The Time-to-Collision metric (TTC) was introduced in 1972 by Hayward~\cite{hayward1972near}. The metric measures the time it takes for an object to collide with another object. The calculation is shown in Eq.~\eqref{eq:ttc_static}-\eqref{eq:ttc_passing}. This widely used metric requires little computational effort and can be applied to different tasks such as motion planning or safety evaluation~\cite{mahmud2017application}. In addition, the metric can be used to determine safety criticality. A road user $V_i$ can be considered a safety-critical object if the TTC between the ego $V_1$ and $V_i$ is less than a defined threshold. However, defining the required threshold can be difficult, as there are different factors that need to be taken into account for safety assessment. Thus, the threshold used in the literature varies between \SIrange{1.0}{4.0}{\second}. 

\paragraph{Stationary Object}
\begin{equation}\label{eq:ttc_static}
    TTC = \frac{d}{|v_1|}
\end{equation}

\paragraph{Passing Collision}
\begin{equation}\label{eq:ttc_passing}
    TTC = \frac{d}{||v_1|-|v_i||}
\end{equation}

\paragraph{Heads-On Collision}
\begin{equation}\label{eq:ttc_heads_on}
    TTC = \frac{d}{||v_1|+|v_i||}
\end{equation}

However, TTC as a stand-alone metric is not suitable for safety assessment. For TTC, only colliding vehicles are considered critical. Consider a motorway situation with a lead vehicle in front of an ego vehicle, both traveling at the same speed and a low distance of only a few meters. The TTC would be equal to $\infty$ and therefore the leading vehicle would not be critical for the perception task. To achieve a safe perception, the vehicle must be perceived and can therefore be seen as safety-critical, which is not indicated by TTC. 

\subsection{Modified Time-to-Collision}
\label{subsubsec:mttc}

A modified version of the TTC (MTTC) as shown in Eq.~\eqref{eq:mttc} was proposed by Ozbay et al.~\cite{ozbay2008derivation} which improves the computability of TTC. The metric itself is extended as Crash Index (CI) (see Eq.~\ref{eq:ci}) with a potential crash severity, which features the ability to be used as a metric for safety assessment. $\Delta v$ and $\Delta a$ describe the relative velocity and acceleration between the two vehicles. 
The metric can be applied for an ego vehicle following a second road user.

\begin{equation}\label{eq:mttc}
    MTTC = \frac{-\Delta v \pm \sqrt{v_1²+2\Delta a_1D}}{\Delta a_1}
\end{equation}

\begin{equation}\label{eq:ci}
\begin{aligned}
    CI = &\frac{(v_1 + a_1 \cdot MTTC)^2 - (v_2 + a_2 \cdot MTTC)^2}{2} \\
    &\cdot \frac{1}{MTTC}
\end{aligned}
\end{equation}

However, the problems of the threshold value and that non-colliding objects are not considered as safety-critical remain.

\subsection{Time-to-Brake}

The Time-to-Brake metric (TTB)~\cite{vanBrummelen2018} represents the time which is left until the ego vehicle must perform a braking maneuver in order to avoid a collision with another vehicle ahead.

\begin{equation}
    TTB = \frac{-(v_{1}-v_{2}) + \sqrt{(v_{1}-v_{2})^2 + 2a_1d_{1,2}}}{a_1}
\end{equation}
 $d_{1,2}$ represents the distance between the two road users. Similar to TTC, the problem of determining an appropriate threshold remains. Hubet et al.~\cite{huber2020evaluation} suggest \SI{1.0}{\second} as threshold.

\subsection{Time-To-Accident}
The Time-to-Accident (TTA) was proposed by Hydén~\cite{hyden1987development}. TTA describes the time to an accident, starting from the moment where one road user starts an evasive action. 
It is defined in~\cite{mahmud2017application} as

\begin{equation}\label{eq:tta}
    TTA = 1.5\cdot \frac{v_1}{16.7 \cdot exp(-0.0306\cdot 0.5V_m)}
\end{equation}

where $V_m$ represents the mean speed. The metric is easily calculated and applicable to the same types of accident scenarios as the TTC. However, as with the TTC, there is the issue of determining an appropriate threshold. The literature review of Mahmud et al.~\cite{mahmud2017application} shows that a threshold of \SI{1.5}{\second} is used in different works. Another disadvantage of the criticality metric is that only accidents are considered safety-critical. However, non-accident scenarios can also contain safety-critical situations and objects.

\subsection{LSM: Braking Distance}
\label{subsec:braking_dist}
Gamerdinger et al.~\cite{gamerdinger2024lsm} proposed a safety metric to evaluate the safety of lane detection. Besides the lateral detection accuracy and the scenario type, they consider the required longitudinal detection range which can be seen as criticality metric. They include an additional safety distance of \SI{10}{\percent} of the braking distance which must remain and define the required longitudinal perception range as shown in Eq.~\eqref{eq:d_long}. $a$ describes the maximum possible braking acceleration and $t_{delay}$ the system delay to start the braking maneuver.

\begin{equation}
        \label{eq:d_long}
        d_{long} = 1.1\cdot (v_1 \cdot t_{delay} + \frac{v_1^2}{2a_1})
\end{equation}

The metric is easy to compute and is suitable for lane detection as well as for the detection of static objects. However, for dynamic objects the braking distance of a single vehicle may not be sufficient in order to avoid a collision.

\subsection{Criticality Index Function}
\label{subsec:cif}
The Criticality Index Function (CIF) proposed by Chan~\cite{chan2006defining} aims to include a potential severity in order to define the criticality in combination with the TTC metric. Chan uses the assumptions that a higher collision speed leads to a higher severity and that a longer time increases the possibility to avoid a collision by an evasive maneuver. The CIF is defined as

\begin{equation}
    \label{eq:cif}
    CIF = \frac{v_1^2}{TTC}
\end{equation}

where TTC is the Time-to-Collision (see Sec.~\ref{subsubsec:ttc}). As for other threshold-based metrics, it is necessary to define a threshold. Chan uses 100, as for a TTC of \SI{1}{\second} a velocity of at most \SI{10}{\metre\per\second} is suitable to avoid a collision by an emergency braking maneuver.

\subsection{RSS}
\label{subsubsec:rss}
The ``Responsible-Sensitive Safety'' (RSS) model was presented by Shalev-Shwartz et al.~\cite{RSS} in 2017 with the goal to guarantee safety in automated driving. RSS is an attempt to formalize human judgment in different road scenarios in a proven mathematical sense. In total, the RSS model contains 34 definitions of safety distances, times, and procedural rules.

These rules specify how an automated vehicle should behave and provide a mathematical description of safe behavior. However, the RSS models do not include metrics to determine safety.

In terms of criticality metrics, the RSS defines different safety distances to be maintained. There are longitudinal and lateral safety distances, which are used by Volk et al.~\cite{volk2020safety} for their perception safety metric.
The longitudinal case is divided into same direction and opposite direction.
With $[x]_+ \coloneqq \max(x,0)$, the three implemented RSS safety distances ($d_s$) are defined as~\cite{RSS}:

 \paragraph{Safety distance longitudinal, same direction} $d_{long,s}$ describes the safety distance lengthwise with the same direction of movement~\cite[Def.1]{RSS}. To determine the required safety distance between the ego $V_1$ and a leading vehicle $V_2$, the respective longitudinal velocities and a response time $\rho$ is considered. $a_{2,\max,\mathrm{brake}}$ is the braking acceleration of $V_2$, while $a_{1,\max,\mathrm{accel}}$ describes the acceleration of $V_1$ during $\rho$. After the response time $V_1$ will decelerate with $a_{1, \min,\mathrm{brake}}$.
\begin{equation}
\begin{split}
d_s= \biggl\lbrack &v_1 \rho + \frac{1}{2} a_{1, \max,\mathrm{accel}} \rho^2  \\
&+\frac{(v_1 + \rho a_{1,\max,\mathrm{accel}})^2}{2a_{1,\min,\mathrm{brake}}} -\frac{v^2_2}{2a_{2,\max,\mathrm{brake}}} \biggl\rbrack _+
\end{split}
\end{equation}

\paragraph{Safety distance longitudinal, opposite direction} $d_{long,o}$ describes the safety distance lengthwise with opposite direction of movement of two road users $V_1$ and $V_2$~\cite[Def. 2]{RSS}. $v_1 \geq 0$ and $v_2 < 0$ describe the longitudinal velocity in relation to the direction of the lane. With $v_{i,\rho} = v_i + \rho a_{i,\max,\mathrm{accel}}$ $d_{long,o}$ is calculated as:
\begin{equation}
\begin{split}
d_{long,o}&=\frac{v_1+v_{1,\rho}}{2}\rho
+\frac{v^2_{1,\rho}}{2a_{1,\min,\mathrm{brake, correct}}}\\
&+\frac{|v_2|+v_{2,\rho}}{2}\rho+\frac{v^2_{2,\rho}}{2a_{2,\min,\mathrm{brake}}}
\end{split}
\end{equation}

$a_{\min,\mathrm{brake, correct}}$ and $a_{\min,\mathrm{brake}}$ describe the braking acceleration of $V_1$ and $V_2$ respectively.

\paragraph{Safety distance lateral}
$d_{lat}$ describes the lateral safety distance between two road users $V_1$ and $V_2$ where $V_1$ is driving to the left of $V_2$. $d_{lat}$ is independent to direction of movement~\cite[Def. 6]{RSS}. It holds $v_{1,\rho} = v_1 + \rho a_{1,\max,\mathrm{accel}}^{lat}$ and $v_{2,\rho} = v_2 - \rho a_{2,\max,\mathrm{accel}}^{lat}$ where $v_1, v_2$ describes the initial lateral speed. $\mu$ is the minimal lateral distance after the two vehicles apply a maximal lateral acceleration towards each other during the time interval $[0, \rho]$ and then apply lateral braking with $a^{\textrm{lat}}_{i,\min,\mathrm{brake}}$ until reaching a lateral speed of 0.
\begin{equation}
\begin{split}
d_{lat}=\mu + \biggl\lbrack &\frac{v_1+v_{1,\rho}}{2}\rho +\frac{v^2_{1,\rho}}{2a^{\textrm{lat}}_{1,\min,\mathrm{brake}}}\\
&-\biggl(\frac{v_2+v_{2,\rho}}{2}\rho
-\frac{v^2_{2,\rho}}{2a^{\textrm{lat}}_{2,\min,\mathrm{brake}}}\biggr) \biggl\rbrack _+
\end{split}
\end{equation}

The RSS model has the advantage that the determination of criticality based on safety distances does not require threshold values and it is easy to determine which metric is required as it is based on the relative position and orientation of the two vehicles.
\subsection{SACRED}
\label{subsec:sacred}
Mori et al.~\cite{mori2023conservative} proposed with Structured Analysis for Conservative Relevance Estimation in Driving context (SACRED) a set of perceptual relevance metrics for motorway scenarios based on German traffic regulations. In total, four different distances are defined, three for radial cases and one for tangential cases such as a merging scenario. The critical distance is always calculated between two road users $V_1$ and $V_2$. The main purpose of their metrics is to reduce the number of vehicles to be considered for trajectory planning. With this goal, they achieved a reduction of about \SI{10}{\percent}.
The SACRED metric is well suited for specific scenarios on motorways, as each of its submetrics is designed for exactly one specific case. However, for the safety evaluation of object detection, the distinction between radial and tangential cases can be difficult in transition between those scenarios, which limits the applicability of the metric as using the wrong metric could falsify the result. For a detailed derivation of the formulae see~\cite{mori2023conservative}.

\paragraph{R.TA: Same Direction (Equal Speed)}
The first relevance metric is for the case of an ego vehicle following a leading vehicle with equal speed. It is defined in~\cite[Eq.16]{mori2023conservative} as:
\begin{equation} \label{eq:FollowingDistance}
\begin{aligned}
    0 < d_\mathrm{min} 
    = & \  d_\mathrm{1,2} - s_\mathrm{1} - s_\mathrm{2} + \frac{v_\mathrm{2,r,0}^2}{2a_{2,\mathrm{max}}}  - v_\mathrm{1,r,0} \ t_\mathrm{1,r} \\
    &- \frac{1}{2} a_\mathrm{max}  t_\mathrm{1,r}^2 
    - \frac{(v_\mathrm{1,r,0} + t_\mathrm{1,r} a_{1,\mathrm{max})^2}}{2 a_\mathrm{1,r,b}}
\end{aligned}
\end{equation} 
where $t_{1,r}$ describes the ego reaction time and $d_{1,2}$ the initial distance between $V_1$ and $V_2$. $v_{i,r,0}$ describes the initial radial velocity. $a_{i,r,b}$ is the radial braking acceleration and $a_{i,\mathrm{max}}$ is the worst case acceleration.
\newpage
\paragraph{R.AT-: Same Direction (Different Speed)}
The second metric of SACRED is for an ego vehicle with a lower speed than desired as for a lane change onto a faster lane, on which a second car is driving. It is defined as~\cite[Eq.21]{mori2023conservative}:
\begin{equation} \label{eq:DistanceMinRAT-}
\begin{aligned}
    0 < d_\mathrm{min} 
    = & \  d(t = t_\mathrm{d}) + \frac{v_\mathrm{1,r,d}^2}{2a_{1,\mathrm{max}}} 
    - v_\mathrm{2,r,d} \ t_\mathrm{2,r} \\
    & - \frac{1}{2} a_{2,\mathrm{max}}  t_\mathrm{2,r}^2 
    - \frac{(v_\mathrm{2,r,d} + t_\mathrm{2,r} a_{2,\mathrm{max}})^2}{2 a_\mathrm{2,r,b}}
\end{aligned}
\end{equation} 
where $d(t = t_\mathrm{d})$ describes the distance between $V_1$ and $V_2$ at the time the ego reaches the desired speed. $v_{i,r,d}$ describes the desired radial velocity while $a_{i,r,d}$ stands for the radial braking acceleration. $a_{max}$ is equal as for Eq.~\eqref{eq:FollowingDistance} and $t_{2,r}$ is in this case the reaction time of the second road user.
\paragraph{R.TT: Opposite Direction}
The third radial case describes the scenario where $V_1$ and $V_2$ move laterally towards each other on opposite sides of the road. For this scenario the minimum distance used as criticality metric is defined as~\cite[Eq.26]{mori2023conservative}:
\begin{equation} \label{eq:DistanceRTT}
\begin{aligned}
    0 < d_\mathrm{min} = 
    & \ d_\mathrm{1,2} - s_\mathrm{1} - s_\mathrm{2} - v_\mathrm{1,r,0} \ t_\mathrm{1,r} \\
    & - \frac{1}{2} a_{1,\mathrm{max}}  t_\mathrm{1,r}^2 - \frac{(v_\mathrm{1,r,0} + t_\mathrm{1,r} a_{1,\mathrm{max}})^2}{2 a_\mathrm{1,r,b}} \\
    & - v_\mathrm{2,r,0} t_\mathrm{1,b} - \frac{1}{2} a_{1,\mathrm{max}} t_\mathrm{1,b}
\end{aligned}
\end{equation}

The variables used are the same as for the same-direction, same-speed case (see Eq.~\eqref{eq:FollowingDistance}) except for the ego braking time $t_{\mathrm{1,b}}$. However, for the safety evaluation of object detection, the applicability of this metric strongly depends on the country. On German motorways, opposing lanes are separated by infrastructural crash barriers to avoid collisions with vehicles coming from the opposite direction. In addition, these infrastructure elements limit the view of the object on the other side; therefore, if the lanes are separated by guardrails, walls or other infrastructure elements, the objects should not be considered for the safety assessment. Therefore, this metric is only relevant for countries with motorways without structural separation between driving directions.

\paragraph{T.XT: Tangential Case (Merging)}
The tangential case is employed in merging scenarios on motorways where an ego attempts to merge in front of $V_2$. Mori et al. defined the required minimal distance as~\cite[Eq.47]{mori2023conservative}:
\begin{equation} \label{eq:DistanceTXT}
\begin{aligned}
    0 < d_\mathrm{min} 
    = & \  d(t = t'_\mathrm{d}) + \frac{v_\mathrm{1,\parallel,d}^2}{2a_{1,\mathrm{max}}} 
    - v'_\mathrm{2,d} \ t_\mathrm{r} \\
    & - \frac{1}{2} a_{1,\mathrm{max}}  t_\mathrm{r}^2 
    - \frac{(v'_\mathrm{2,d} + t_\mathrm{r} a_{2,\mathrm{max})^2}}{2 a_\mathrm{1,\parallel,g}}
\end{aligned}
\end{equation} 
where $t_r$ and $a_{max}$ are equal to the radial cases. Similar to Eq.~\ref{eq:DistanceMinRAT-} $d(t = t_\mathrm{d})$ describes the distance between $V_1$ and $V_2$ at the time the ego reaches the desired speed. 
$v_{2,d}' = v_{2,0}+a_{max}t'_d$ stands for the desired velocity of $V_2$ under assumption of a maximal acceleration. $v_\mathrm{1,\parallel,d}$ as lateral desired velocity is assumed to be equal to the initial velocity of $V_2$ ($v_{2,0}$). $a_\mathrm{1,\parallel,g}$ represents the lateral acceleration of $V_1$.

\subsection{SURE-Val}
\label{subsec:urban_darmstadt}
Storms et al.~\cite{storms2023sure} extended the work from \cite{mori2023conservative} by adapting one of the radial and tangential metrics to urban scenarios. However, the specific use case of urban scenarios limits the applicability for safety assessment in different scenarios, as a set of different metrics would be required, which could falsify the criticality classification. For a detailed derivation of the formulae see~\cite{storms2023sure}.

\paragraph{Opposite Direction with Static Obstacle}
This metric is an adapted version of Eq.~\eqref{eq:DistanceRTT} by Mori et al.~\cite{mori2023conservative}.
It's adapted to the case of a two lane road with $V_1$ on a road with a static obstacle and a road user $V_2$ driving on the other lane in opposite direction. In contrast to the motorway scenario in~\cite{mori2023conservative}, here no separation between the driving directions is present. The required safety distance is defined in~\cite[Eq.12]{storms2023sure} as:
\begin{equation} \label{eq:DistanceRTT'}
\begin{aligned}
    0 < d_\mathrm{min} = 
    & \ (d_\mathrm{1,2} - d_\mathrm{1,e} - d_\mathrm{2,e}) - s_\mathrm{1} - s_\mathrm{2} - v_\mathrm{1,r,a} \ t_\mathrm{1,r} \\
    & - \frac{1}{2} a_\mathrm{max}  t_\mathrm{1,r}^2 - \frac{(v_\mathrm{1,r,a} + t_\mathrm{1,r} a_{1,\mathrm{max}})^2}{2 a_\mathrm{1,r,b}} \\
    & - v_\mathrm{2,r,e} t_\mathrm{1,b} - \frac{1}{2} a_{2,\mathrm{max}} t_\mathrm{1,b}
\end{aligned}
\end{equation}
with $s_1,s_2$ being the lateral dimension of $V_1$ and $V_2$, respectively. The distances $d_{1,2}$ are the distance between the two vehicles, the total movement distance of the ego and the acceleration distance of $V_2$. $a_{max}$ describes the worst case acceleration as for SACRED~\cite{mori2023conservative}. $v_i$ stands for the velocity of car $c_i$, the same holds for $a_i$. $t_{1,r}$ denoted the reaction time and $t_{1,b}$ the braking maneuver time of $V_1$. However, the dataset can only be applied to the case in which $V_1$ must evade a static obstacle. This scenario is not present in any dataset which is suitable for the evaluation; hence, this metric cannot be evaluated and is only presented for the sake of completeness.
\paragraph{T.XT' Tantential Case (Crossing)}
For intersection scenarios an adapted version of the tangential metric (T.XT) by Mori et al.~\cite{mori2023conservative} (see Eq.~\eqref{eq:DistanceTXT}) is proposed.
Instead of the merging on motorway, in this case a car $V_1$ and another car $V_2$ are driving on an intersection with crossing trajectories. The minimum distance (here denoted as $r_\mathrm{1,\bot,0}$) is defined as~\cite[Eq.16]{storms2023sure}:
\begin{equation}
\begin{aligned}
    r_\mathrm{1,\bot,0} & < 
    v_\mathrm{1,\bot,0} t_\mathrm{r} + \sqrt{\frac{v_\mathrm{1,\bot,0}^2}{2 a_\mathrm{1,\bot,b}}}\\
    & +
    \frac{1}{2} a_{1,\mathrm{max}} \left[t_\mathrm{1,r} + \frac{v_\mathrm{1,\bot,0} + t_\mathrm{1,r} a_{2,\mathrm{max}} }{a_\mathrm{1,b}} \right]^2 
\end{aligned}
\end{equation}
The variable notation is equal to Eq.~\eqref{eq:DistanceRTT'}.
\subsection{Futher Metrics}
\label{subsec:different_metrics}
As shown by Mahmud et al.~\cite{mahmud2017application} and Westhofen et al.~\cite{westhofen2023criticality}, there are many other metrics. However, some of them, such as Time-to-Zebra~\cite{varhelyi1998drivers}, are not suitable for evaluating object detection. Other metrics such as Steer Threat Number or Break Threat Number by Jannson~\cite{jansson2005collision} can be used as a metric to evaluate an object detection system; however, they do not define any criteria for the need to perceive an object and are therefore not suitable as a criticality metric in terms of perception safety evaluation. The Time-to-Steer metric (TTS)~\cite{vanBrummelen2018} can be considered as criticality metric; however, in many situations steering is not a possible action due to limited drivable area and further objects. Also determining, if a steering action is possible before applying the metric can be considered as complex. Hence, this metric can be considered as not suitable. In addition, the safety metric by Volk et al.~\cite{volk2020safety} is a safety metric that uses the RSS metric (see Sec.~\ref{subsubsec:rss}) as a criticality metric, but does not provide any further relevance criteria.

\section{APPLICATION STRATEGIES}
\label{sec:strategies}
As the results for the single metrics show, here is potential to improve the state-of-the art criticality metrics (see Sec.~\ref{subsec:results}). Since the novel metrics can be complex in application, we provide two approaches called: multi-metric aggregation and a bidirectional criticality rating.
\subsection{Multi-Metric Aggregation}
\label{subsec:combination}
Different metrics provide different strengths and weaknesses. In addition, generalizing criticality over all scenarios is a highly complex problem which could lead to incomprehensible results. Moreover, proving the meaningfulness and correctness of such metrics could be difficult, which limits their suitability. Therefore, we propose the idea to aggregate different metrics to overcome the limitations of single metrics. If one of the metrics classifies the other vehicle as critical, it is marked as critical. To compensate the weakness of a metric, the combination of metrics should consider different aspects such as time and distance. Using correlating metrics such as TTC and CIF are not likely to increase the performance.

Hence, we consider the following combinations:
\begin{itemize}
    \item TTC + (RSS/LSM/SACRED/SURE-VAL)
    \item TTA + (RSS/LSM/SACRED/SURE-VAL)
\end{itemize}

TTC and TTA are combined with one of RSS, LSM, SACRED or SURE-VAL. Using CIF or MTTC is not considered as they are correlated to TTC and showed lower results. 

\subsection{Bidirectional Criticality Rating}
\label{subsec:bidirectional}
Criticality metrics are not necessarily inverse. This means if a metric $M$ applied on vehicle $V_1$ does not classify a vehicle $V_2$ as critical, the same metric applied to $V_2$ could classify $V_1$ as critical. Considering safety evaluation, in this case a critical situation between $V_1$ and $V_2$ could occur and therefore $V_2$ must be considered as critical for $V_1$ even though $M$ classified $V_2$ as uncritical. 
Hence, we suggest to perform a bidirectional criticality rating which can be formulated as
\begin{equation}
    \begin{aligned}
    C(V_i)=&(M(V_e \rightarrow V_i)=\text{True}) || \\
    &(M(V_i \rightarrow V_e)=\text{True})
\end{aligned}    
\end{equation}

where $V_e$ represents the ego vehicle, $C(V_i)$ the criticality of $V_i$ and $M(V_i\rightarrow V_j)$ the application of the metric $M$ from $V_i$ to $V_j$. This bidirectional criticality rating is performed $\forall V_i \in \{V \setminus V_e\}$.
\section{EVALUATION}
\label{sec:eval}
In order to ascertain the suitability of criticality metrics for the safety evaluation of object detection systems, it is necessary to consider safety-critical situations, such as accidents. General accident datasets frequently comprise solely of information such as the time of day, driver age, and so forth, as their principal objective is the prediction of accident severity, with the consequence that information pertinent to safety evaluation, such as vehicle dimensions, velocities, and object classes, is frequently absent. Wang et al.~\cite{Wang_2023_DeepAccident} proposed the Deep Accident, a diverse dataset of 57k frames designed for accident prediction and collective perception. The dataset comprises approximately 690 scenarios, encompassing both accident and non-accident scenarios. 
The evaluation of criticality metrics is facilitated by all scenarios that incorporate a safety-critical situation, defined in this case by an accident. Given that the dataset only provides complete meta information within the train data, 159 scenarios with a total of 4995 frames were extracted from the train set, incorporating the IDs of colliding objects. 
In each scenario two vehicles, one denoted as ego and a second object collide. For the evaluation, the metrics are applied to determine if the other object is considered as safety-critical from the view of the ego vehicle.
As the goal of this review is to determine the suitability of criticality metrics for the offline safety evaluation of object detection systems, we use the ground truth data provided in the dataset as input for the metrics. 

\begin{figure*}
\centering

\begin{subfigure}[b]{0.85\textwidth}
   \includegraphics[width=\linewidth, trim= 0cm 0.2cm 0cm 0.0cm, clip]{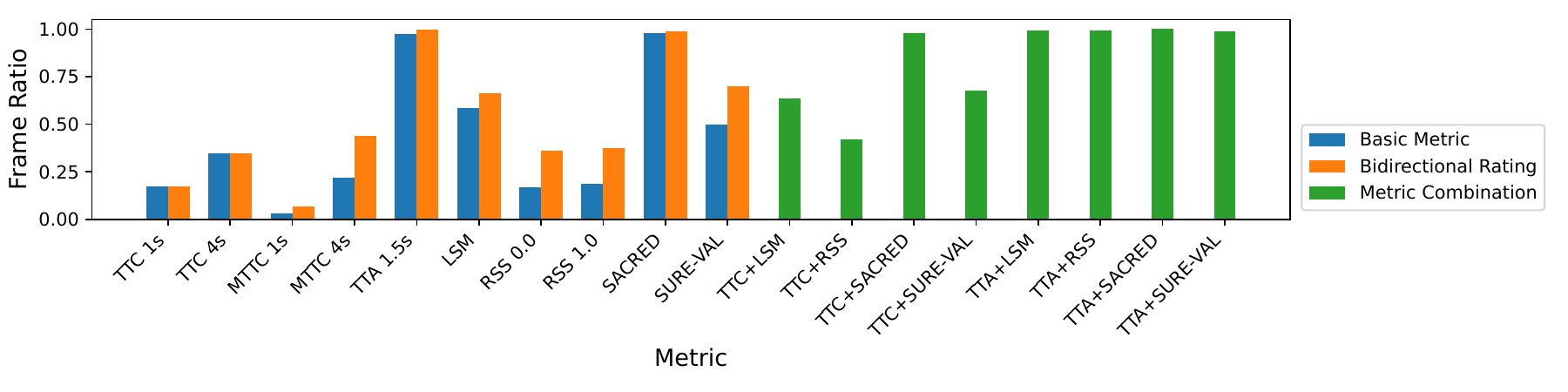}
   \caption{Frame Ratio Results}
   \label{fig:results_fr}
\end{subfigure}

\begin{subfigure}[b]{0.85\textwidth}
   \includegraphics[width=\linewidth, trim= 0cm 0.2cm 0cm 0.0cm, clip]{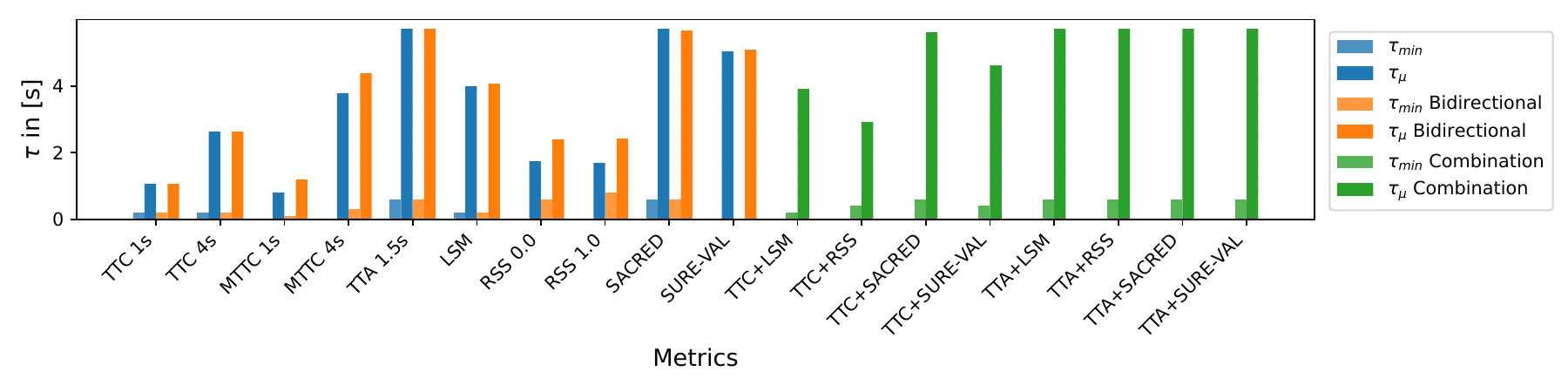}
   \caption{$\tau$ Results}
   \label{fig:results_tau}
\end{subfigure}
\caption{Results for Frame Ratio and $\tau$, Basic Metrics in Comparison to Bidirectional and Combination Approach}
\label{fig:results}
\end{figure*}

As metrics to evaluate the criticality metrics in this study, we employ the scenario ratio $\mathrm{SR}$ which describes the ratio of scenarios in which at least one classification of the other object as safety-critical is present. Furthermore, the ratio of frames $\mathrm{FR}$ in which the other object is classified as safety-critical, as well as the time between the first marking of the other object as safety-critical and the collision $\tau$, are considered in the evaluation. Results are presented in Tab.~\ref{tab:basic_metrics}-~\ref{tab:bidirectional}. A graphical excerpt of results for a better comparison between basic metrics and our optimization strategies is shown in Fig.~\ref{fig:results}.

\subsection{Results for Original Metrics}
\label{subsec:results}
In order to enable a meaningful statement about the criticality metrics, for threshold-based metrics different literature-based thresholds are applied. The used thresholds are based on the conducted literature review by Mahmud et al.~\cite{mahmud2017application}. 
For TTC and MTTC we report results with  $T$ from \SIrange{1}{4}{\second}, for TTB and TTA the threshold is set to \SI{1.0}{\second} and \SI{1.5}{\second}, respectively. For CIF we use thresholds of \{80, 100, 120\}. For the drift margin $\mu$ of the RSS lateral safety distance we use \{0.0, 0.1, 0.5, 1.0\}\si{\metre}.

\begin{table}[t]
      \caption{Results for single metric evaluation. First Basic Metrics (see Sec.~\ref{subsubsec:ttc}-~\ref{subsec:cif}), then RSS (see Sec.~\ref{subsubsec:rss}) and SACRED as well as SURE-VAL (see Sec.~\ref{subsec:sacred}-~\ref{subsec:urban_darmstadt}). $\tau$ (mean, std, min)}
      \centering
    \begin{tabular}{ll    rrr} \toprule
    Metric &$T$ / $\mu$&$\mathrm{SR}$& $\mathrm{FR}$ & $\tau$\\ \midrule
    TTC& \SI{1}{\second}&0.977& 0.171 & (1.055, 0.413, 0.2)  \\
    TTC& \SI{2}{\second}&0.977& 0.268 & (1.902, 0.875, 0.2)  \\
    TTC& \SI{3}{\second}&0.977& 0.321 & (2.343, 1.141, 0.2)  \\
    TTC& \SI{4}{\second}&0.977& 0.346 & (2.620, 1.377, 0.2)  \\
    MTTC& \SI{1}{\second}&0.721& 0.032 & (0.785, 1.317, 0.0)  \\
    MTTC& \SI{2}{\second}&0.895& 0.086 & (1.644, 1.512, 0.0) \\
    MTTC& \SI{3}{\second}&0.953& 0.152 & (2.696, 1.708, 0.0)  \\
    MTTC& \SI{4}{\second}&0.977& 0.219 & (3.775, 1.928, 0.0)  \\
    TTB&\SI{1.0}{\second}&\textbf{1.000}& 0.749 & \textbf{(5.708, 1.728, 0.6})\\
    TTA&\SI{1.5}{\second}&\textbf{1.000}& \textbf{0.976} & \textbf{(5.708, 1.728, 0.6)}\\
    CIF& 80&0.977& 0.146 & (0.868, 0.565, 0.0)  \\
    CIF& 100&0.977& 0.127 & (0.739, 0.459, 0.0)  \\
    CIF& 120&0.977& 0.107 & (0.638, 0.414, 0.0)  \\
    LSM& - &0.953& 0.583 & (3.979, 1.371, 0.2) \\\midrule
    \midrule
    RSS&0.0&0.814 & 0.167 & (1.724, 1.260, 0.0)  \\
    RSS&0.1& 0.826 &0.169 & (1.706, 1.263, 0.0)  \\
    RSS&0.5& 0.872&0.179 & (1.644, 1.265, 0.0)  \\
    RSS&1.0&0.895 &0.188 & (1.668, 1.239, 0.0)  \\\midrule
    \midrule
    R.TA& - &0.477 & 0.039 & (1.037, 1.818, 0.0)  \\
    R.AT& -&0.512 &0.073 & (2.084, 2.892, 0.0)\\
    R.TT& -& 0.988&0.852 & \textbf{(5.418, 1.621, 1.0)}  \\
    R.AA& -&0.163 &0.005 & (2.521, 3.139, 0.0) \\
    T.XT& -&0.023 &0.0004 & (4.300, 3.000, 1.3)  \\
    T.XT'& -&0.756 &0.498 & (5.023, 2.348, 0.0)  \\
    \midrule
    SACRED& -& \textbf{1.000} & \textbf{0.979} & \textbf{(5.708, 1.728, 0.6)}  \\
    SURE-VAL & -&0.756 &0.498 & (5.023, 2.348, 0.0)  \\
    
    \bottomrule
    \bottomrule
    
\end{tabular}
\label{tab:basic_metrics}
\end{table}
Table~\ref{tab:basic_metrics} shows the results of the evaluation of all single metrics. All basic metrics achieved similar results for $\mathrm{SR}$ ranging between 0.90 and 1.00, meaning that in up to \SI{100}{\percent} of the scenarios, the colliding object is identified as safety-critical for at least one frame. Only for MTTC \SI{1}{\second} a lower score of 0.72 was achieved. For the frame ratio $\mathrm{FR}$, the results range between 0.107 and 0.976. Here, the CIF and MTTC performed worst, while the LSM, TTB and TTA classified the other vehicle correctly as safety-critical over \SI{58}{\percent} of the frames. Considering the time between the first classification as safety-critical and the accident, again the TTA and LSM performed really well with a \SI{5.7}{\second} and \SI{3.9}{\second} respectively. However, for nearly all metrics it can be observed that $\tau_{min}=\SI{0.0}{\second}$, meaning that just in the frame of the accident the metric predicts the object as safety-critical. Only TTA and TTB achieved better results with $\tau_{min}=\SI{0.6}{\second}$.

As for the basic metrics the$\mathrm{SR}$ results for RSS are similar over all threshold variations and ranges between 0.81 and 0.89. The frame ratio corresponds to the scenario ratio values and range between 0.16 and about 0.19 where a higher $\mathrm{SR}$ value comes along a higher $\mathrm{FR}$ value. For $\tau$ also the results over all drift margins are similar and are about \SI{1.7}{\second} in average. Furthermore, equal to the basic metrics $\tau_{min}$ is \SI{0.0}{\second}.

The SACRED and SURE-VAL line represent the combination of the single submetrics. In contrast to the basic metrics and the RSS model a much higher range in $\mathrm{SR}$ is present. The values range between 0.023 for T.XT and about 0.99 for R.TT. The $\mathrm{FR}$ values correlate with the $\mathrm{SR}$ values, with higher $\mathrm{FR}$ values for higher $\mathrm{SR}$ results. However, it can be seen that the $\mathrm{FR}$ values are much lower than the $\mathrm{SR}$ values, showing that only for a few frames per scenario the classification works properly. Considering the $\tau$ results, SACRED and SURE-Val achieved high results with \SI{1.04}{\second} for R.TA and up to \SI{5.42}{\second} for R.TT. Showing that the colliding object is classified as safety-critical a sufficient before they collide. Remarkable are the $\tau_{min}$ results for R.TT and T.XT, showing that if a correct prediction is made, it's also a sufficient time before the collision. 

Since the radial metrics and the T.XT metric belong together, they are evaluated in combination as SACRED leading to remarkable results with $\mathrm{SR}=1.0$ and $\mathrm{FR}=0.979$. In addition, for $\tau$ the result could be increased to a minimum time of \SI{0.6}{\second}. Furthermore, SURE-VAL is listed as an extra metric; however, as the SURE-VAL scenario of a parking vehicle is not included in the dataset the SURE-VAL result corresponds only to the T.XT' metric.

\subsection{Results for Application Strategies}
\label{subsec:results2}

Results for the multi-metric aggregation approach are shown in Tab.~\ref{tab:combination}. For TTC and RSS, the threshold with the best performance from the single metric evaluation is taken. For TTC in combination with LSM, $\mathrm{SR}$ can be slightly increased, but $\mathrm{FR}$ can be increased from about 0.3 to 0.637. For RSS, similar results can be observed with a slight increase in $\mathrm{FR}$. For TTC and TTA combined with SACRED, the $\mathrm{SR}=1.0$ is from the TTA metric, but as for LSM and RSS, a significant increase in $\mathrm{FR}$ of up to \SI{100}{\percent} to about 0.99 can be observed.
\begin{table}[H]
      \caption{Results for Multi-Metric Aggregation (see Sec.~\ref{subsec:combination}), $\tau$ (mean, std, min)}
      \centering
    \begin{tabular}{ll    rrr} \toprule
    Metric& $T/\mu$ &$\mathrm{SR}$& $\mathrm{FR}$ & $\tau$\\ \midrule
    TTC+LSM&\SI{4}{\second} / -&0.988 & 0.637 & (3.904, 1.408, 0.2)  \\
    TTC+RSS&\SI{4}{\second} / 1.0&0.977 & 0.418 & (2.915, 1.147, 0.4)  \\
    TTC+SACRED&\SI{4}{\second} / -&1.00 & 0.979 & (5.606, 1.664, 0.6)  \\
    TTC+SURE-VAL&\SI{4}{\second} / -&1.00 & 0.677 & (4.616, 2.175, 0.4)  \\
    TTA+LSM&\SI{1.5}{\second} / -&1.00 & 0.995 & (5.708, 1.728, 0.6)  \\
    TTA+RSS&\SI{1.5}{\second} / 1.0&1.00 & 0.993 & (5.708, 1.728, 0.6)  \\
    TTA+SACRED&\SI{1.5}{\second} / -&1.00 & 1.00 & (5.708, 1.728, 0.6)  \\
    TTA+SURE-VAL&\SI{1.5}{\second} / -&1.00 & 0.988 & (5.708, 1.728, 0.6)  \\\bottomrule
    \bottomrule
\end{tabular}
\label{tab:combination}
\end{table}

Table~\ref{tab:bidirectional} shows the results for the bidirectional criticality rating. For TTC, MTTC and RSS the lowest and highest threshold values are evaluated to demonstrate the improvement.

Looking at the TTC and MTTC results, it can be observed that the scenario ratio does not increase compared to the base results; however, the $\mathrm{FR}$ results for MTTC are increased by about \SI{100}{\percent}. Also the $\tau$ results could be increased by up to \SI{5.71}{\second} as the mean for MTTC \SI{4}{\second}, especially the minimum value could be increased from \SI{0.0}{\second} to \SI{0.3}{\second}. For TTA, which already had an SR of 1.0 in the base results, the $\mathrm{FR}$ could be increased by 0.02 using the bidirectional evaluation. A similar behavior can be observed for LSM, where $\mathrm{SR}$ does not change, but $\mathrm{FR}$ and $\tau$ show a slight increase.
\begin{table}[H]
      \caption{Results for Bidirectional Criticality Rating (see Sec.~\ref{subsec:bidirectional}), $\tau$ (mean, std, min)}
      \centering
    \begin{tabular}{ll    rrr} \toprule
    Metric& $T/\mu$ &$\mathrm{SR}$& $\mathrm{FR}$ & $\tau$\\ \midrule
    TTC&\SI{1}{\second}&0.977 & 0.171 & (1.055, 0.413, 0.2)  \\
    TTC&\SI{4}{\second}&0.977 & 0.346 & (2.620, 1.377, 0.2)  \\
    MTTC&\SI{1}{\second}&0.721 & 0.066 & (1.179, 1.368, 0.1)  \\
    MTTC&\SI{4}{\second}&0.977 & 0.438 & (4.367, 1.831, 0.3)  \\
    TTA&\SI{1.5}{\second}&1.00 & 0.996 & (5.708, 1.728, 0.6)  \\
    LSM& - &0.953 & 0.662 & (4.048, 1.349, 0.2)  \\
    RSS& 0.0 &0.977 & 0.359 & (2.383, 0.992, 0.6)  \\
    RSS& 1.0 &0.977 & 0.375 & (2.417, 0.985, 0.8)  \\
    SACRED& - &1.00 & 0.990& (5.660, 1.688, 0.6)  \\
    SURE-VAL& - &0.942 & 0.699 & (5.091, 2.360, 0.0)  \\
    \bottomrule
\end{tabular}
\vspace*{5mm}
\label{tab:bidirectional}
\end{table}
For the RSS metric, the highest increase can be observed. For $\mu = 0.0$ the $\mathrm{SR}$ can be increased from 0.814 to 0.977, also $\mathrm{FR}$ is increased by about \SI{100}{\percent}. Similar behavior can be observed for $\mu=1.0$ with $\mathrm{SR}=0.977$ and $\mathrm{FR}=0.375$. Particularly, the increase for $\tau$ must be noted, where the minimum time can be increase from \SI{0.0}{\second} to \SI{0.6}{\second} and \SI{0.8}{\second} for $\mu=0.0$ and $\mu=1.0$, respectively.

SACRED already showed good results for the basic results, hence, only a slight increase can be achieved with the bidirectional approach. For SURE-VAL, the scenario ratio $\mathrm{SR}$ can be significantly increased from 0.756 to 0.94 and the frame ratio $\mathrm{FR}$ from 0.498 to 0.699. Both corresponds to an increase of about \SI{50}{\percent}. Regarding $\tau$ only a minor increase can be observed.

In total, the bidirectional approach shows an increase in all metrics. Especially, the RSS metric and SACRED can significantly achieve better results by applying the bidirectional approach.
\newpage
\section{DISCUSSION}
\label{sec:discussion}
Each of the metrics analyzed has its advantages and disadvantages. TTC and MTTC are easy to calculate, widely used, and reasonable. Moreover, both achieve good results with up to \SI{97.7}{\percent} correctly classified scenarios. However, their performance is highly dependent on the threshold used. A higher threshold leads to better classification of security-critical objects, but the prediction becomes less accurate. In addition, the actual threshold required is highly dependent on the scenario and the motion planning algorithms used. This is also an issue for the TTA, which achieved the best result of all metrics using a threshold of \SI{1.5}{\second} as proposed by Mahmud et al.~\cite{mahmud2017application}. The CIF metric, which combines speed with TTC, is also easy to calculate and incorporates speed, which is a safety-related factor. However, this metric is rarely used and is less easy to interpret. As the last of the basic metrics, the braking distance used by Gamerdinger et al.~\cite{gamerdinger2024lsm} performed well and achieved similar results in terms of scenario classification. For the frame ratio and the time between the first safety-critical classification and the collision, the metric performed best together with TTA. However, the metric is used in the safety evaluation of lane detection and therefore considers only a single vehicle. Considering and accumulating the braking distances of both vehicles could further improve the result.

The RSS metric is mathematically proven to achieve safety in automated driving. In addition, it is already used in safety assessment as a criticality metric~\cite{volk2020safety}. The metric is comprehensible and includes the safety-relevant factors of acceleration, speed and vehicle dimensions. However, the lateral safety distance requires the definition of a drift margin as a threshold, which may vary depending on the scenario. In addition, the RSS metric performed slightly worse than the basic metrics with $\mathrm{SR}$ values of 0.81 up to 0.89, which is about 0.10 less than TTA. 

The SACRED and SURE-Val metrics achieved good results in terms of object reduction for motion planning. However, they performed worst in our evaluation, with the exception of the R.TT metric, which achieved results similar to the best metrics. The other metrics only achieved $\mathrm{SR}$ values between 0.023 and 0.76, which is significantly lower compared to the other metrics. In addition, the SACRED metrics are designed for application to motorway scenarios and the SURE-Val metrics are designed specifically for urban scenarios. For some scenarios it can also be difficult to correctly identify which sub-metric to apply. 

In summary, all metrics have some weaknesses in generalization or performance. However, the RSS metric, SACRED, and the basic TTA and stopping distance metrics give good results as single metrics and can be considered the most suitable to be used as criticality metrics for the safety evaluation of object detection.

In addition, we have shown that our two approaches, multi-metric aggregation and bidirectional rating, can further increase the results for all metrics. For the bidirectional approach for all metrics, the frame ratio can be increased. For the scenario ratio and $\tau$ especially for RSS and SURE-VAL an increase could be achieved. Using a combination of time and distance based metrics, scenario ratios of $\geq 0.98$ could be achieved with frame ratios of 0.418 - 1.00, which is a significant improvement. These two approaches allow for better generalization to overcome scenario-specific limitations of a single application of a metric. This results in a more meaningful assessment of criticality. Therefore, bidirectional rating or using a combination of metrics should be preferred over a single metric.


\section{CONCLUSION \& OUTLOOK}
\label{sec:conclusion}

In this paper, we presented a comprehensive technical analysis of criticality metrics tailored for safety evaluation in object detection systems of automated vehicles. We conducted an extensive empirical study using the DeepAccident dataset, focusing on safety-critical scenarios. Our findings demonstrate significant limitations in widely used Time-to-X metrics, which depend on predefined thresholds and often yield unreliable or inconsistent results. Additionally, many existing metrics lack generalization, as they are applicable only to narrowly defined scenarios, limiting their effectiveness in real-world safety assessment.

Among the evaluated approaches, TTA, TTB, and SACRED achieved the highest accuracy in correctly identifying critical scenarios. However, even these metrics failed to classify the colliding object as critical in up to \SI{50}{\percent} of the relevant frames. This highlights a key shortcoming: current state-of-the-art criticality metrics are not sufficiently reliable for robust safety evaluation in object detection tasks.

To address these challenges, we introduced two novel approaches, a bidirectional criticality assessment and a multi-metric aggregation strategy. These methods significantly improved the criticality classification accuracy, leading to a relative increase of up to \SI{100}{\percent} in the identification of critical frames, thereby offering a more nuanced and reliable estimation of criticality.

Future work will focus on leveraging these insights to further refine criticality metrics for perception evaluation. In particular, incorporating object-specific dynamics—such as differentiating between vehicle types, pedestrians, or cyclists—may substantially improve the contextual relevance and accuracy of the metrics. Furthermore, we plan to release the implementation developed in this study as an open-source criticality metrics library. This will support broader adoption and reproducibility, enabling the research community to build upon our work in advancing safety assessment methodologies for automated vehicles.






\bibliographystyle{IEEEtran} 
\bibliography{literature.bib}

@article{RSS,
  author    = {Shai Shalev{-}Shwartz and
               Shaked Shammah and
               Amnon Shashua},
  title     = {{On a Formal Model of Safe and Scalable Self-driving Cars}},
  journal={ArXiv},
  volume    = {abs/1708.06374},
  year      = {2017},
}

@article{westhofen2023criticality,
  title={Criticality metrics for automated driving: A review and suitability analysis of the state of the art},
  author={Westhofen, Lukas and Neurohr, Christian and Koopmann, Tjark and Butz, Martin and Sch{\"u}tt, Barbara and Utesch, Fabian and Neurohr, Birte and Gutenkunst, Christian and B{\"o}de, Eckard},
  journal={Archives of Computational Methods in Engineering},
  volume={30},
  number={1},
  year={2023},
  publisher={Springer}
}

@inproceedings{volk2020safety,
  title={A comprehensive safety metric to evaluate perception in autonomous systems},
  author={Volk, Georg and Gamerdinger, J{\"o}rg and von Bernuth, Alexander and Bringmann, Oliver},
  booktitle={2020 IEEE 23rd International Conference on Intelligent Transportation Systems (ITSC)},
  year={2020},
  organization={IEEE}
}

@article{Wang_2023_DeepAccident,
    title = {DeepAccident: A Motion and Accident Prediction Benchmark for V2X Autonomous Driving},
    author = {Wang, Tianqi and Kim, Sukmin and Ji, Wenxuan and Xie, Enze and Ge, Chongjian and Chen, Junsong and Li, Zhenguo and Ping, Luo},
    journal = {arXiv preprint arXiv:2304.01168},
    year = {2023}
}

@inproceedings{mori2023conservative,
  title={Conservative estimation of perception relevance of dynamic objects for safe trajectories in automotive scenarios},
  author={Mori, Ken and Storms, Kai and Peters, Steven},
  booktitle={2023 IEEE International Conference on Mobility, Operations, Services and Technologies (MOST)},
  year={2023},
  organization={IEEE}
}

@article{hayward1972near,
  title={Near miss determination through use of a scale of danger},
  author={Hayward, John C},
  year={1972},
  publisher={Pennsylvania State University University Park}
}

@article{storms2023sure,
  title={Sure-val: Safe urban relevance extension and validation},
  author={Storms, Kai and Mori, Ken and Peters, Steven},
  journal={arXiv preprint arXiv:2308.02266},
  year={2023}
}

@article{ozbay2008derivation,
  title={Derivation and validation of new simulation-based surrogate safety measure},
  author={Ozbay, Kaan and Yang, Hong and Bartin, Bekir and Mudigonda, Sandeep},
  journal={Transportation research record},
  volume={2083},
  number={1},
  pages={105--113},
  year={2008},
  publisher={SAGE Publications Sage CA: Los Angeles, CA}
}

@article{mahmud2017application,
  title={Application of proximal surrogate indicators for safety evaluation: A review of recent developments and research needs},
  author={Mahmud, SM Sohel and Ferreira, Luis and Hoque, Md Shamsul and Tavassoli, Ahmad},
  journal={IATSS research},
  volume={41},
  number={4},
  pages={153--163},
  year={2017},
  publisher={Elsevier}
}

@INPROCEEDINGS{gamerdinger2024lsm,
  author={Gamerdinger, Jörg and Teufel, Sven and Amann, Stephan and Volk, Georg and Bringmann, Oliver},
  booktitle={2024 IEEE 100th Vehicular Technology Conference (VTC2024-Fall)}, 
  title={LSM: A Comprehensive Metric for Assessing the Safety of Lane Detection Systems in Autonomous Driving}, 
  year={2024},
  volume={},
  number={},
  pages={1-7},
  doi={10.1109/VTC2024-Fall63153.2024.10757827}}

@phdthesis{jansson2005collision,
  title={Collision Avoidance Theory: With application to automotive collision mitigation},
  author={Jansson, Jonas},
  year={2005},
  school={Link{\"o}ping University Electronic Press}
}

@inproceedings{chan2006defining,
  title={Defining safety performance measures of driver-assistance systems for intersection left-turn conflicts},
  author={Chan, Ching-Yao},
  booktitle={2006 IEEE Intelligent Vehicles Symposium},
  pages={25--30},
  year={2006},
  organization={IEEE}
}

@article{hyden1987development,
  title={The development of a method for traffic safety evaluation: The Swedish Traffic Conflicts Technique},
  author={Hyd{\'e}n, Christer},
  journal={Bulletin Lund Institute of Technology, Department},
  number={70},
  year={1987}
}

@article{varhelyi1998drivers,
  title={Drivers' speed behaviour at a zebra crossing: a case study},
  author={Varhelyi, Andras},
  journal={Accident Analysis \& Prevention},
  volume={30},
  number={6},
  year={1998},
  publisher={Elsevier}
}

@article{vanBrummelen2018,
  author = {van Brummelen, Jeroen and O'Brien, Maarten and Gruyer, Dennis and Najjaran, Hugo},
  title = {Autonomous Vehicle Control: A Review of Decision Making Techniques and Their Implications for Autonomous Driving},
  journal = {IEEE Transactions on Intelligent Transportation Systems},
  volume = {PP},
  number = {99},
  pages = {1--14},
  year = {2018},
  doi = {10.1109/TITS.2018.2836307},
  publisher = {IEEE}
}

@inproceedings{huber2020evaluation,
  title={Evaluation of virtual traffic situations for testing automated driving functions based on multidimensional criticality analysis},
  author={Huber, Bernd and Herzog, Steffen and Sippl, Christoph and German, Reinhard and Djanatliev, Anatoli},
  booktitle={2020 IEEE 23rd International Conference on Intelligent Transportation Systems (ITSC)},
  pages={1--7},
  year={2020},
  organization={IEEE}
}

\end{document}